\documentclass[10pt,twocolumn,letterpaper]{article}

\usepackage[pagenumbers]{wacv} 

\newcommand{\red}[1]{{#1}}
\newcommand{\blue}[1]{{#1}}

\definecolor{wacvblue}{rgb}{0.21,0.49,0.74}
\usepackage[pagebackref,breaklinks,colorlinks,allcolors=wacvblue]{hyperref}
\usepackage{algorithm}
\usepackage{algpseudocode}
\usepackage{multirow}
\def\wacvPaperID{665} 
\def\confName{WACV}
\def\confYear{2026}

\title{HDR Reconstruction Boosting with \\ Training-Free and Exposure-Consistent Diffusion}

\author{
Yo-Tin Lin$^{1}$\quad
Su-Kai Chen$^{2}$\quad
Hou-Ning Hu$^{2}$\quad
Yen-Yu Lin$^{1}$\quad
Yu-Lun Liu$^{1}$\vspace{0.25em}
\\
\centerline{$^1$National Yang Ming Chiao Tung University \quad $^2$MediaTek Inc.}\vspace{0.25em}
\\
 {\url{https://github.com/EusdenLin/HDR-Reconstruction-Boosting}}
}
\begin{document}

\twocolumn[{%
\renewcommand\twocolumn[1][]{#1}%
\maketitle
\begin{center}
\centering
\captionsetup{type=figure}
\resizebox{1.0\textwidth}{!} 
{
    \includegraphics[width=\textwidth]{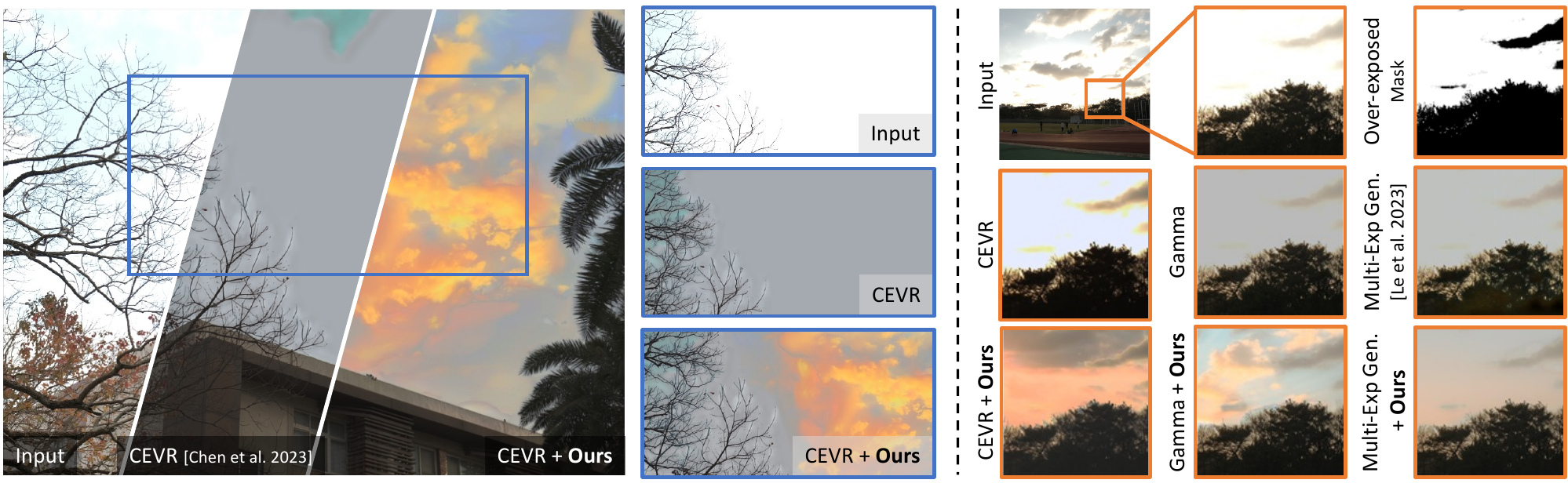}
    }
    \vspace{-6mm}
    \caption{
    \textbf{Our training-free, diffusion-based approach enhances existing HDR reconstruction methods for over-exposed regions.} (\emph{Left}) Our method improves CEVR~\cite{chen2023learning} by generating natural sky details in the over-exposed area. (\emph{Right}) Comparisons with different baseline methods (CEVR, SingleHDR~\cite{liu2020single}, and Multi-Exposure Generation~\cite{le2023single}) demonstrate how our approach consistently improves reconstruction quality across different methods without requiring additional training. Here are the results of the \emph{real-world} cases captured by a Fujifilm X-T30 DSLR camera.
    }
 	\label{fig:teaser}
\end{center}
}]
\maketitle
\begin{abstract}
Single LDR to HDR reconstruction remains challenging for over-exposed regions where traditional methods often fail due to complete information loss. We present a training-free approach that enhances existing indirect and direct HDR reconstruction methods through diffusion-based inpainting. Our method combines text-guided diffusion models with SDEdit refinement to generate plausible content in over-exposed areas while maintaining consistency across multi-exposure LDR images. Unlike previous approaches requiring extensive training, our method seamlessly integrates with existing HDR reconstruction techniques through an iterative compensation mechanism that ensures luminance coherence across multiple exposures. We demonstrate significant improvements in both perceptual quality and quantitative metrics on standard HDR datasets and in-the-wild captures. Results show that our method effectively recovers natural details in challenging scenarios while preserving the advantages of existing HDR reconstruction pipelines.
Project page: \href{https://github.com/EusdenLin/HDR-Reconstruction-Boosting}{here}.
\end{abstract}    
\vspace{-4mm}
\section{Introduction}
\label{sec:intro}

High Dynamic Range (HDR) imaging aims to capture and reproduce the wide spectrum of luminance levels present in real-world scenes. This encompasses both extremely bright and very dark regions, which typically exceed the dynamic range of conventional imaging sensors and standard displays. To address this, HDR reconstruction methods have been developed to recover information lost during the signal processing pipeline of Low Dynamic Range (LDR) images. However, as illustrated in Fig.~\ref{fig:teaser}, while existing methods such as CEVR~\cite{chen2023learning} excel at reconstructing correctly exposed regions, they often struggle with over-exposed areas due to the limited information in these regions.

\begin{figure}[t]
\centering
\includegraphics[width=1.0\columnwidth]{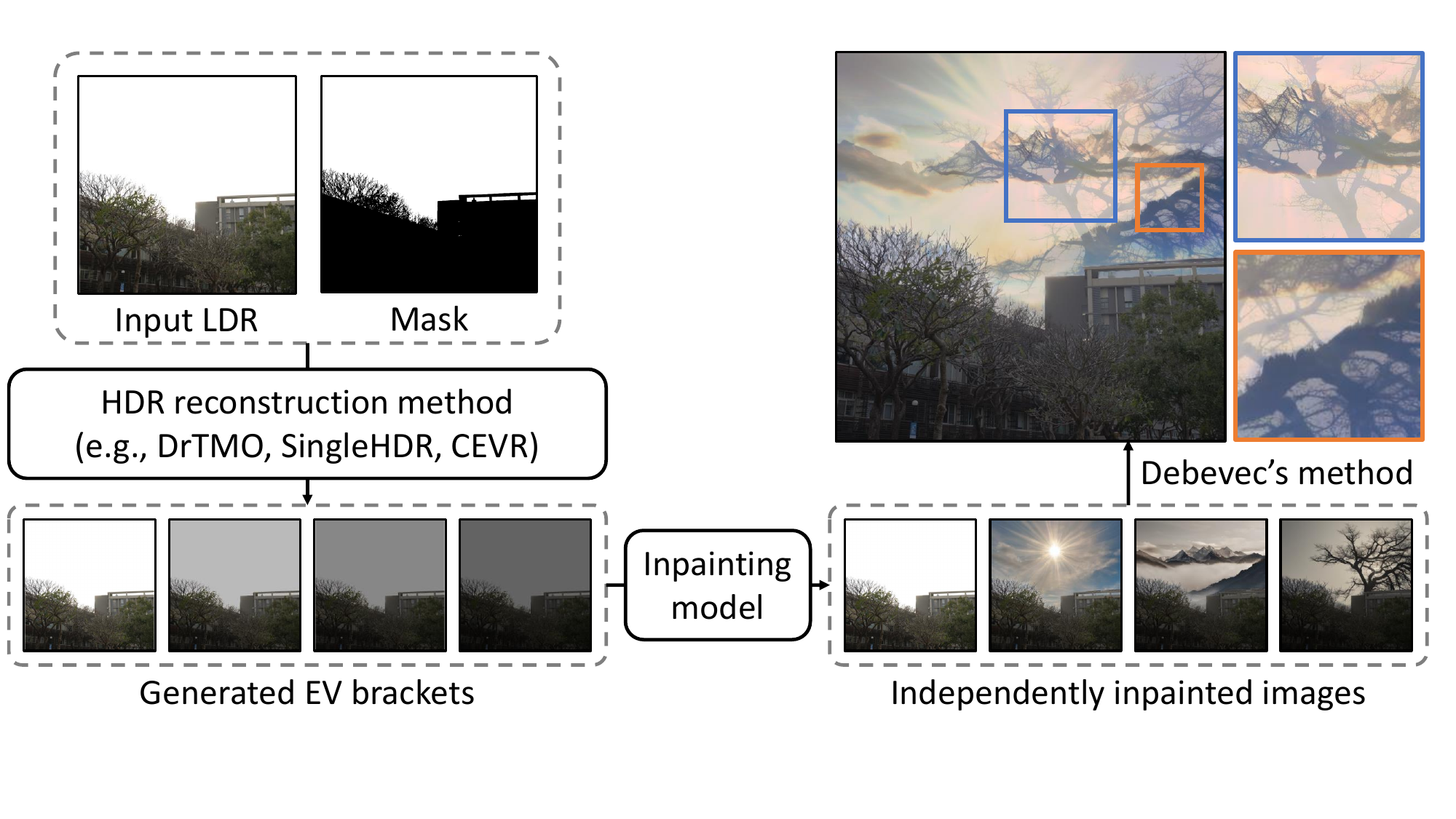}
\vspace{-12mm}
\caption{
\textbf{Limitations arising from naively combining indirect HDR reconstruction methods and over-exposed regions inpainting.} 
%
Independent inpainting of each EV bracket, without cross-EV alignment, can introduce ghosting artifacts in the merged HDR result. 
These artifacts stem from inconsistencies in the generated content, becoming apparent after merging the independently inpainted exposures using Debevec's method.
%
}
\label{fig:alignment}
\vspace{-0.15in}
\end{figure}

Over-exposed regions pose a significant challenge in enhancing image quality and realism. These regions, characterized by the complete loss of texture and detail, undermine the visual coherence and fidelity of reconstructed images. Conventional methods, including those based on tone mapping curves or learning-based approaches, are often inadequate as evidenced by the gamma correction results in Fig.~\ref{fig:teaser}. These regions typically appear as uniform patches devoid of texture, making accurate reconstruction an ill-posed problem. Even advanced GAN-based methods, such as \cite{wang2023glowgan, lee2018deep}, despite their promise, suffer from generalization issues and inconsistent reconstruction quality, particularly in severe over-exposed scenarios. Moreover, as illustrated in Fig.~\ref{fig:alignment}, direct application of existing diffusion inpainting models to independently inpaint different exposures often leads to artifacts such as ghosting and color inconsistencies due to misalignment across exposure levels and inadequate control of inter-texture consistency.

In this work, we propose a generative method that leverages diffusion priors to address the limitations of existing HDR reconstruction techniques. Our method employs an image inpainting paradigm to hallucinate plausible content in over-exposed regions, prioritizing perceptual quality and natural transitions over strict adherence to ground truth. As demonstrated in Fig.~\ref{fig:teaser} (\emph{Right}), our approach can significantly enhance existing methods like CEVR and Multi-Exposure Generation by generating natural and consistent details in over-exposed areas.

To achieve this, we incorporate HDR reconstruction through multi-exposure image synthesis. By aligning and compensating for discrepancies across low dynamic range (LDR) images with different exposure values(EVs), we ensure consistency and luminance accuracy in the generated HDR image. Our pipeline includes three key components: (1) {\em a diffusion-based inpainting backbone} to ensure high-quality generation tailored to over-exposed regions, (2) {\em iterative refinement with SDEdit \cite{meng2021sdedit}} to maintain luminance consistency and structural coherence and to gradually enhance the generated content, and (3) {\em compensation after every inpainting step} to make the generated content exceed the expected luminance lower bounds, avoiding unrealistic brightness or structural artifacts.

In summary, the key contributions of this work include: 
\begin{itemize}
    \item Introducing a diffusion-based inpainting pipeline that effectively hallucinates over-exposed regions, significantly enhancing visual quality and aesthetic appeal.
    \item Proposing a robust iterative compensation and inpainting strategy using SDEdit to align luminance and maintain texture consistency across LDR in the brackets.
    \item Showing the versatility of our approach, which seamlessly integrates with existing indirect and direct HDR reconstruction methods without additional training.
\end{itemize}

\begin{figure*}[t]
\centering
\includegraphics[width=1.0\textwidth]{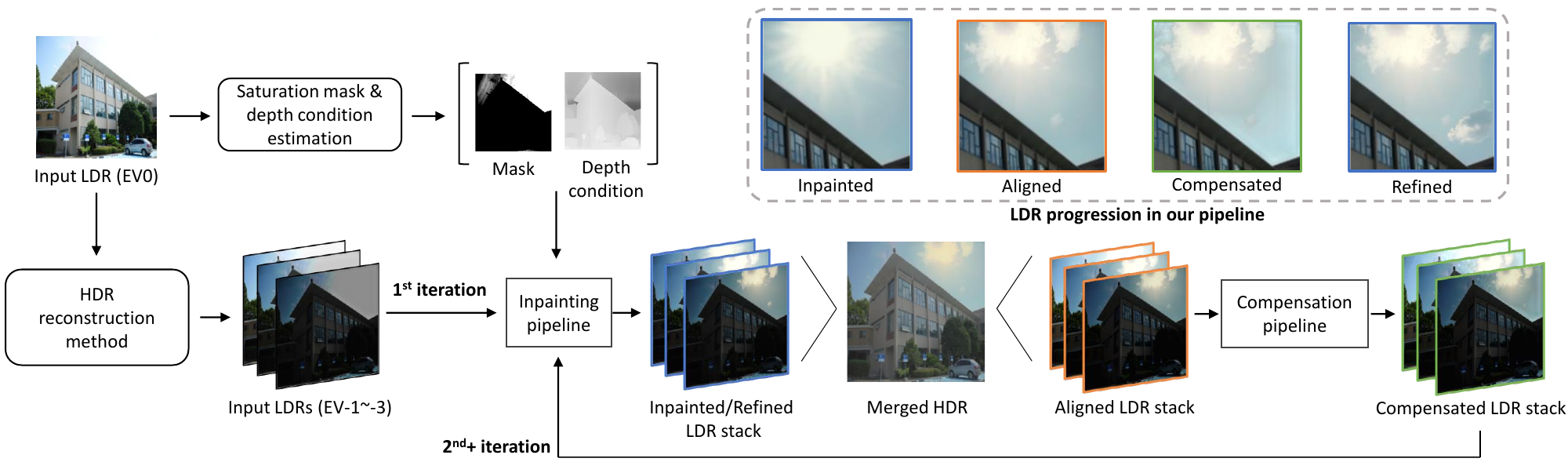}
\vspace{-8mm}
\caption{
\textbf{Overview of our training-free HDR reconstruction pipeline.} 
Given an input LDR image (EV0), we generate bracketed LDR images using an existing HDR reconstruction method. 
Our iterative pipeline then enhances these results through (1) an inpainting stage guided by exposure and condition maps, (2) HDR merging and alignment of the generated content, and (3) a compensation stage to ensure physical consistency. 
The top-right inset shows the progressive refinement of the over-exposed regions through our pipeline stages.
The upper sky region in the aligned case shows the clear alignment across EV images. 
The lower sky region in the compensation and refined cases demonstrates our algorithm's ability to hallucinate realistic over-expose regions with plausible intensity and texture.
%
The upper part of the sky region in the aligned case shows the clear alignment across EV images. The lower parts of the sky region in the compensation and refined cases show the ability of our algorithm to hallucinate realistic over-expose regions with reasonable intensity and texture.
%
}
\label{fig:overview}
\vspace{-0.15in}
\end{figure*}

\section{Related Work}
\paragraph{HDR reconstruction.}
Single-image HDR methods are either direct or indirect. Direct methods, including HDRCNN~\cite{eilertsen2017hdr}, ExpandNet~\cite{marnerides2018expandnet}, \red{CNN models~\cite{khan2019FHDR,lee2022HSVNET,wu2022LiTMNet}}, masked features~\cite{santos2020single}, and pipeline reversal~\cite{liu2020single}, predict HDR directly but struggle with ill-posed over-exposed regions. Indirect methods merge generated multi-exposure LDR images via inverse response estimation~\cite{debevec2023recovering,robertson2003estimation,mitsunaga1999radiometric}. While methods like CEVR \cite{chen2023learning}, Deep Chain~\cite{endo2017deep,lee2018deep}, and Recursive HDRI~\cite{lee2018re} excel in proper exposures, they often fail in saturated areas. Recent approaches explore cycle-consistency~\cite{lee2020learning}, exposure control~\cite{jo2021deep}, differentiable learning~\cite{kim2021end}, generative models~\cite{zhang2021deep}, GANs~\cite{wang2023glowgan}, and decomposition~\cite{Sebastian2024Intrinsic}. Our method improves the indirect approach using diffusion priors~\cite{dhariwal2021diffusion,ho2020denoising} to reconstruct over-exposed content while retaining exposure bracketing benefits.

\vspace{-3mm}
\paragraph{Over-exposed region handling for HDR.}
Reconstructing over-exposed regions remains challenging due to information loss~\cite{eilertsen2017hdr, santos2020single, zhang2021deep, liu2020single}. While GAN-based methods like GlowGAN~\cite{wang2023glowgan} (using StyleGAN-XL~\cite{sauer2022stylegan}) exist, they often lack generalization or require scene-specific training~\cite{lee2018re, niu2021hdr, zhang2021deep}. In contrast, diffusion approaches like Exposure Diffusion~\cite{bemana2024exposure}, DiffusionLight~\cite{phongthawee2024diffusionlight}, and LEDiff~\cite{wang2024lediff} generalize well across scenes~\cite{dhariwal2021diffusion, rombach2022high, nichol2021glide}. Similarly, LightsOut~\cite{tsai2025lightsout} employs diffusion to reconstruct lost light sources. Leveraging diffusion's synthesis~\cite{lugmayr2022repaint, avrahami2022blended} and control capabilities~\cite{ho2020denoising, song2020denoising}, our method achieves diverse generation without scene-specific training.

\vspace{-3mm}
\paragraph{Diffusion \& image inpainting.}
Recent diffusion-based inpainting methods, such as Blended Latent Diffusion~\cite{avrahami2023blended}, use CLIP guidance~\cite{radford2021learning} and noise blending for smooth transitions~\cite{rombach2022high, nichol2021improved}. Several studies explore controlled generation~\cite{dhariwal2021diffusion, ho2020denoising}, such as RePaint~\cite{lugmayr2022repaint}, ControlNet~\cite{zhang2023adding}, SDEdit~\cite{meng2021sdedit}, and IP-Adapter~\cite{ye2023ip}. Paint-by-Example~\cite{yang2023paint} shows effective CLIP-guided inpainting, while others explore personalization~\cite{tang2024realfill} and specialized architectures~\cite{preechakul2022diffusion, wallace2023edict} to enhance inpainting capabilities. To improve generation fidelity, CorrFill~\cite{liu2025corrfill} introduces correspondence guidance for reference-based inpainting. Furthermore, recent works have expanded diffusion priors to traditional regression tasks, such as matting~\cite{wang2024matting}.
Unlike methods requiring scene-specific training~\cite{wang2023glowgan, niu2021hdr, zhang2021deep} or fine-tuning~\cite{phongthawee2024diffusionlight}, our training-free approach achieves high-quality results by controlling pre-trained diffusion models to maintain cross-exposure consistency.
\section{Method}

Given an LDR image containing over-exposed regions, our objective is to reconstruct an HDR image, with particular emphasis on restoring details within the over-exposed areas.
Using existing indirect HDR reconstruction methods (Section 3.2), we can generate predicted LDR images with varying exposure value (EV) offsets (\eg, EV: -1, -2, -3) and obtain the estimated inverse camera response function (CRF) using Debevec's method \cite{debevec2023recovering}.
After applying inpainting techniques to these images, we can create low-EV images containing reconstructed over-exposed details, which are utilized to synthesize the final HDR image.



\begin{figure*}[t]
\centering
\includegraphics[width=1.0\textwidth]{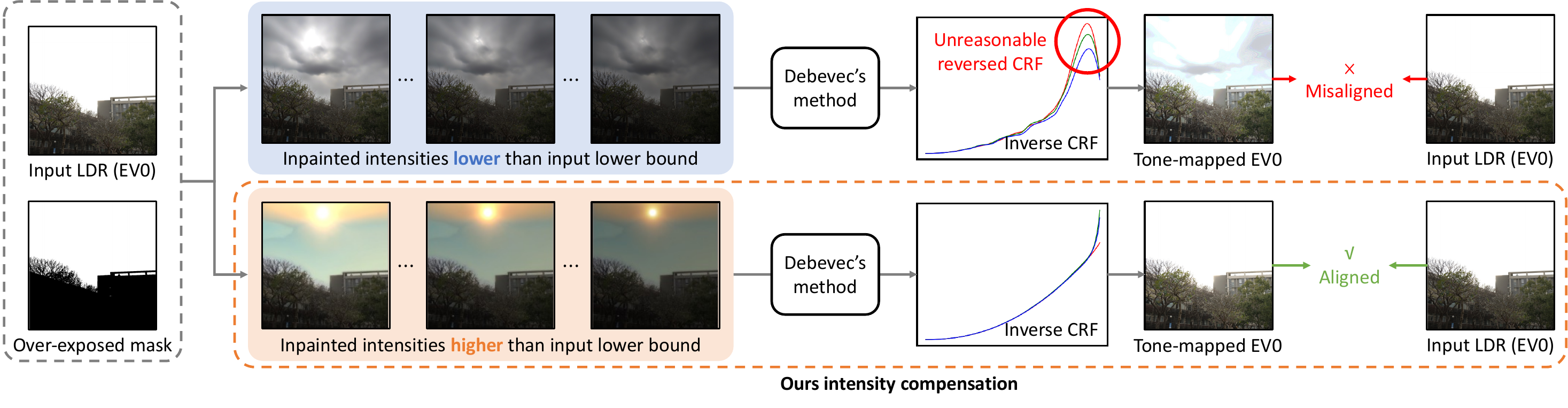}
\vspace{-8mm}
\caption{
\textbf{Motivation for luminance compensation in HDR reconstruction.} The top row shows the existence of inpainted intensities below the lower bound of over-exposed regions may lead to unreasonable inverse CRF estimation and misaligned results. 
The bottom row demonstrates our intensity compensation approach, which ensures bounded intensities, resulting in proper inverse CRF estimation and exposure-aligned reconstruction that matches the input LDR.}
\label{fig:compensation}
\vspace{-0.15in}
\end{figure*}

However, as shown in Fig.~\ref{fig:compensation}, if the generated details violate the lower-bound constraint, \ie, the minimum luminance level in over-exposed regions, the inverse CRF predicted by reconstruction methods can be disrupted.
This disruption leads to color shifts and other artifacts, causing the reconstructed HDR image to deviate from the original LDR input. 
Furthermore, while diffusion models can produce consistent textures with varying brightness levels under a fixed random seed, the iterative nature of inpainting may result in a loss of cross-EV consistency. This leads to chaotic results in the final HDR image, as shown in Fig.~\ref{fig:alignment}. 


To address the challenges of lower-bound adherence and cross-EV alignment, we propose a novel inpainting framework comprising two key components: the Inpainting Pipeline (Section 3.3) and the Compensation Pipeline (Section 3.4). 
An overview of our algorithm is shown in Fig.~\ref{fig:overview}. 
Our method employs an iterative inpainting algorithm with a single LDR image as input. 
At each iteration, we refine the saturated areas using a corresponding text prompt, ensuring that the inpainted pixels satisfy the lower-bound constraint and remain consistent across different EV levels.


\subsection{Preliminaries}
\paragraph{ControlNet \cite{zhang2023adding}.}
It provides a mechanism to condition the generation process on additional structural guidance, such as edges or depth maps. 
A standard latent diffusion model (LDM) operates through a denoising process: $z_\text{t-1} = f_{\theta}(z_t, t, y)$, where $z_t$ is the noisy latent variable at timestep $t$, and $y$ is the text prompt. 
The function $f_\theta$ represents the U-Net denoiser parameterized by $\theta$. 
ControlNet introduces an external conditioning signal $c$, modifying the generation process to $z_\text{t-1} = f_{\theta}(z_t, t, y, c)$, where $c$ ensures that the generated image preserves structural consistency. 
Using ControlNet prevents the model from generating unrealistic textures in over-exposed regions, making the generated content align with the physical scene structure.


\vspace{-0.15in}
\paragraph{Inpainting Models.}

To combine ControlNet with the inpainting diffusion process \cite{avrahami2022blended}, we use a conditioning mechanism to ensure that the inpainted region adheres to both structural constraints and textual guidelines. 
The modified denoising function is
%
\begin{equation}
    z_\text{t-1} = f_{\theta}(z_t, t, y, c) \odot (1-M) + \hat{\text{X}}_\text{t} \odot M,
\end{equation}
where $M$ is the saturation mask, $\hat{\text{X}}_\text{t}$ is the target image at timestep $t$ in the forward process, $c$ is the ControlNet guidance (depth map), and $\odot$ is the element-wise multiplication.
 

\subsection{LDR Preprocess}

Given an LDR image, our objective at this stage is to estimate its appearance under different exposure values, which is essential to CRF estimation. 
\blue{To this end, we utilize an existing indirect or direct HDR reconstruction method to generate LDR images across varying exposure values (EVs).}


For cases where a direct HDR reconstruction method, such as \cite{liu2020single}, is preferred, we can generate the baseline LDR stacks using the following gamma function:
%
%
\begin{equation}
{X}_\text{EV} = (H\times2^\text{EV})^{\gamma},
\end{equation}
where $X_{EV}$ represents the EV-adjusted image of the normalized HDR input $H$ at the exposure level $0$ (EV0). 
We set $\gamma$ to $2.2$ in our implementation and apply this gamma function to all direct methods in the experiments.


Using the generated LDR stacks, we estimate the inverse CRF via Debevec's method~\cite{debevec2023recovering}. 
This estimated function serves as the reference for both tone-mapping and HDR merging, and is also used within our compensation pipeline (Section 3.4).


Additionally, we identify saturated regions in the image by using the soft mask formula \cite{eilertsen2017hdr}:
%
\begin{equation}
    m_i = \frac{\max(0, \max_c(X_{i, c}-\tau))}{255-\tau},
\end{equation}
where $X$ is the input image with pixel index $i$ and channel index $c$, and $X_{i,c} \in [0,255]$. 
In our implementation, we set the threshold $\tau$ to $245$. 
To mitigate edge side effects caused by iterative inpainting, we apply this soft mask, marking saturated areas for the first iteration, to reduce edge artifacts. 


\begin{figure}[t]
\centering
\includegraphics[width=1.0\columnwidth]{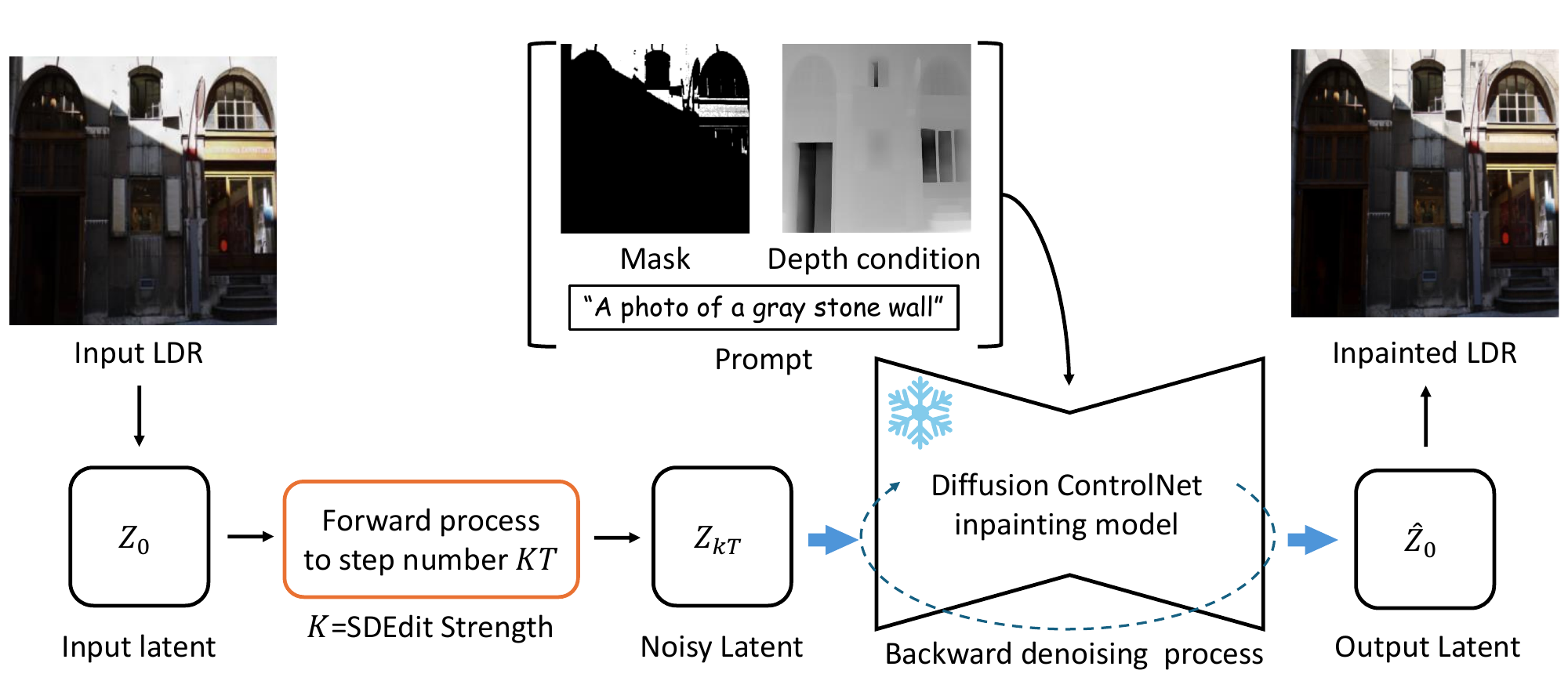}
\vspace{-8mm}
\caption{
\textbf{Inpainting pipeline in our method.} 
The depth-conditioned inpainting pipeline with scheduled strength is able to generate reasonable and consistent content across different EVs and iterations.
The scheduled SDEdit strength enables us to balance detail preservation with creative generation. In the early iterations, we allow the model to explore a broader range of plausible scene details, while in later iterations, we retain the refined details from previous steps and only update pixels that do not meet the physical constraints.
}
\vspace{-0.15in}
\label{fig:inpainting_pipeline}
\end{figure}

\subsection{Inpainting Pipeline}

A primary challenge during inpainting is that blended diffusion inpainting models~\cite{avrahami2023blended} often disregard the pixel values generated by baseline methods in over-exposed regions. 
They tend to produce visually appealing but brightness-inconsistent results.


To address this issue, we employ SDEdit~\cite{meng2021sdedit}. Unlike the standard forward process, which adds noise up to the full timestep $T$, we stop at a partial timestep $t$, where $t<T$. 
Inspired by the observation that brightness information is primarily determined in early timesteps, we set $t$ to the range $[0.85T, 0.95T]$ in our implementation.
Furthermore, we find that smaller values of $t$ typically result in closer alignment with the results in the previous iteration.
Therefore, we develop a scheduler that starts at $0.95T$ and then gradually reduces the value of $t$ throughout the iterative inpainting process. 
This mechanism helps balance the diversity of generated results while maintaining texture consistency across iterations.

As depicted in Fig.~\ref{fig:inpainting_pipeline}, we use the standard diffusion inpainting model as the backbone model. 
The proposed inpainting pipeline takes a prompt, the soft mask, and the depth map as the extra guidance, and can generate more plausible LDR images for HDR reconstruction.


\begin{figure}[t]
\centering
\includegraphics[width=1.0\columnwidth]{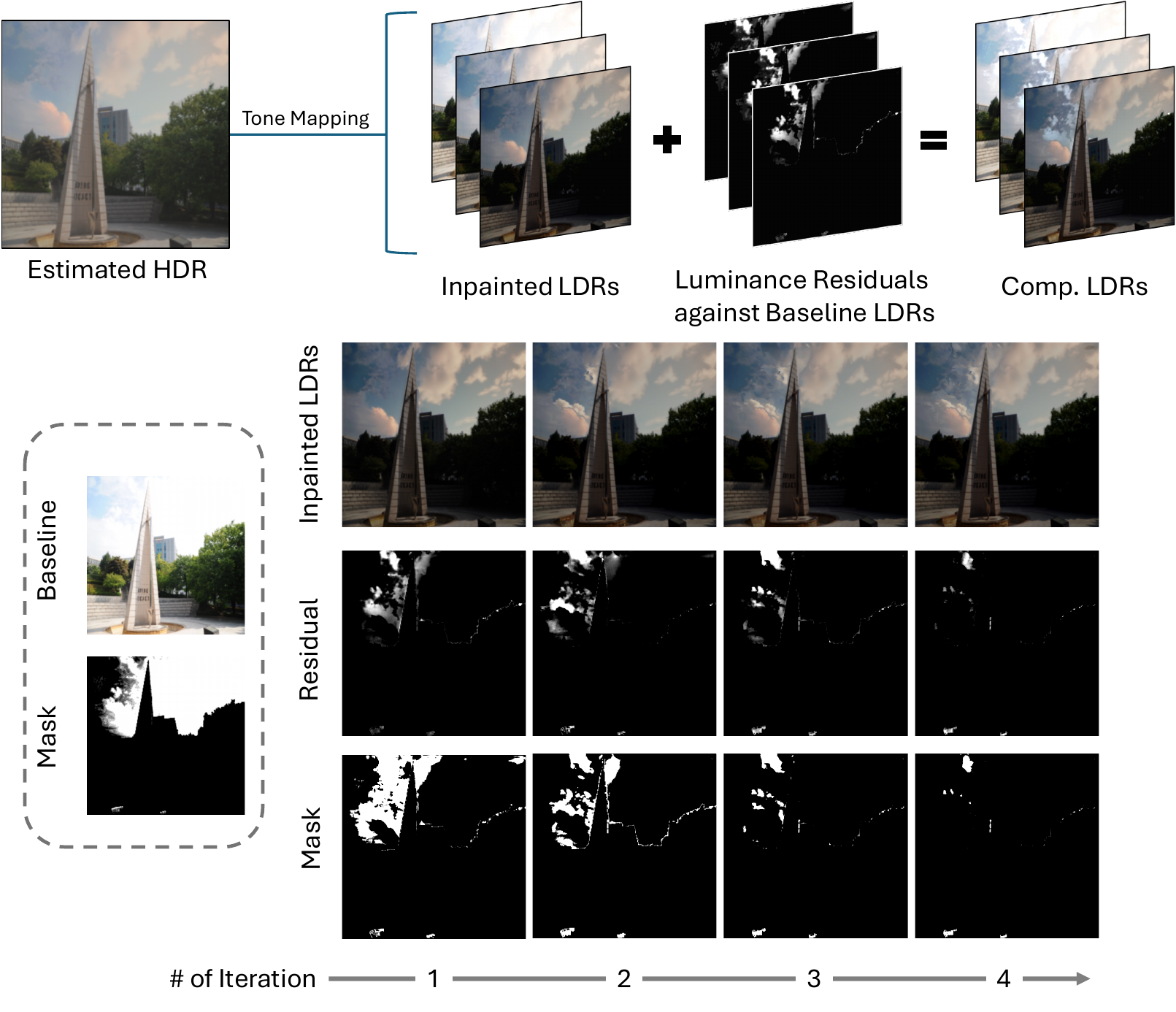}
\vspace{-8mm}
\caption{
\textbf{Compensation pipeline in our method.} Our pipeline ensures proper exposure relationships through iterative refinement. (\emph{Top}) Overview of how inpainted LDRs are combined with luminance residuals to produce compensated results after tone-mapping. (\emph{Bottom}) Visualization over four iterations shows inpainted results, decreasing residuals, and shrinking mask regions. The process maintains proper luminance lower bounds while focusing refinement on problematic areas through selective masking, preventing CRF estimation issues.
}
\label{fig:compensation_pipeline}
\end{figure}

\subsection{Compensation Pipeline}

As shown in Fig.~\ref{fig:compensation}, inpainting with intensities below the input's lower bound leads to inaccurate inverse CRF estimation and misaligned results. 
When the generated content fails to maintain consistent luminance relationships across different exposures, it disrupts the inverse CRF predicted by reconstruction methods and causes artifacts in the final output. 
To address this issue, we introduce a compensation pipeline that ensures cross-EV consistency and proper intensity lower bounds.


To maintain cross-EV consistency among LDR images throughout the iterative process, we utilize the inverse CRF estimated by baseline methods: The LDR stacks are first merged into an HDR image \cite{Banterle2017HDRtoolbox}, which is then converted back into LDR stacks with matching EV levels, using the same CRF.


After obtaining the aligned LDR stacks, we ensure that the generated regions meet the desired lower-bound constraints. 
To achieve this, we transform both the generated image and the baseline image into the YUV space to extract luminance. 
By calculating the residual between the baseline and the generated image and multiplying it by a compensation scale, we derive the values to adjust the generated image. 
These adjusted values are passed to the next iteration for refinement.


To prevent abrupt brightness increases that could cause artifacts and disrupt smooth transitions between adjacent regions, the compensation scale is initially set to a low value and gradually increases with each iteration.


To ensure that valid generated regions remain unchanged in subsequent iterations, as shown in Fig.~\ref{fig:compensation_pipeline}, we create a mask during the residual calculation. 
Valid pixels are assigned a mask value of $0$, while invalid pixels are set to $1$. 
This shrinks the inpainting region, allowing the model to focus specifically on areas that do not meet the standard.
As illustrated in the figure, the compensation pipeline effectively reduces the residual, ultimately ensuring that all pixels meet the expected criteria. 
Alg.~\ref{algorithm:overview} summarizes the process from the LDR stacks to the final HDR results.


\begin{algorithm}[t]
\small
	\caption{Inpainting Pipeline}\label{algorithm:overview}
	\begin{algorithmic}[0]
            \State \textbf{Input:} Baseline LDR images $X=\{X_{0}, X_{-1}, X_{-2}, X_{-3}\}$, text prompt $p$, total step number $T$, condition image $c$, SDEdit strength $K$, compensation scale $S$, saturation mask $m$, total inpainting iterations $N$,  estimated inverse CRF $\textit{inv.-CRF}$, exposure times $times$
            \State \textbf{Output:} HDR image $H$
            \State $\hat{X}\leftarrow X$
            \For {$i=1,2,\ldots, N$}
			\For {$EV=-1,-2,-3$}
                    \State $Z_\text{EV, t} = \texttt{noise}(\hat{X}_\text{EV}, t)$
                    \For {\textbf{all} \textit{t} from $T \times K$ to \textit{1}}
				    \State $Z_\text{EV, t-1, fg} = \texttt{noise}(\hat{X}_\text{EV}, t-1)$
                        \State $Z_\text{EV, t-1, bg} = \texttt{denoise}(Z_{EV, t}, c, p)$
                        \State $Z_\text{EV, t-1} = Z_\text{EV, t, fg} \odot m + Z_\text{EV, t, bg} \odot (1-m))$
                    \EndFor 
                \State $\hat{X}_{\text{EV}, \text{inpaint}} = Z_\text{EV, 0}$
			\EndFor
		\State $H = \texttt{merge}(\hat{X}_\text{inpaint}, \textit{inv.-CRF}, times)$
            \State $\hat{X}_\text{align} = \texttt{tonemap}(H, \textit{inv.-CRF}, times)$
		\State $\text{residual} = X - \hat{X}_\text{align}$
            \State $\hat{X}_\text{comp} = \hat{X}_\text{align} + \text{residual} \times m \times S_i$
            \State $m\leftarrow m \times (\text{residual} > 0)$
            \State $\hat{X} \leftarrow \hat{X}_\text{comp}$
		\EndFor
        \State return $H$
	\end{algorithmic} 
\end{algorithm}



\section{Experiments}

\begin{table*}[t]
\centering
\caption{\textbf{Quantitative comparisons.} \blue{Our algorithm consistently improves state-of-the-art methods on both VDS~\cite{lee2018deep} and HDR-Eye~\cite{nemoto2015visual} datasets without additional training.} }
\vspace{-3mm}
\label{tab:baseline_ours}
\resizebox{\textwidth}{!}{%
\begin{tabular}{l|l*{14}{c}}
\toprule 

 & & \multicolumn{11}{c}{Non reference} & \multicolumn{3}{c}{Ref.-based (vs. GT)} \\
\cmidrule(lr){3-13} \cmidrule(lr){14-16}
Dataset & Method & \multicolumn{2}{c}{BRISQUE ↓} & \multicolumn{2}{c}{NIQE ↓} & \multicolumn{2}{c}{NIMA ↑} & \multicolumn{2}{c}{MUSIQ ↑} & \multicolumn{2}{c}{CLIP-IQA ↑} & PU21 ↓ & \multicolumn{2}{c}{KID ↓} & HDR- ↑ \\
& & & & & & & & & & & & \multicolumn{1}{l}{PIQE} & & & \multicolumn{1}{l}{VDP-3} \\
\cmidrule(lr){3-4} \cmidrule(lr){5-6} \cmidrule(lr){7-8} \cmidrule(lr){9-10} \cmidrule(lr){11-12} \cmidrule(lr){14-15}
& \multicolumn{1}{r}{TMO} & RH's & KK's & RH's & KK's & RH's & KK's & RH's & KK's & RH's & KK's & -- & RH's & KK's & -- \\
\midrule
\multirow{9}{*}{VDS} & SingleHDR~\cite{liu2020single} & 75.73 & 73.08 & 14.49 & 14.25 & 4.021 & 4.352 & 2.032 & 0.295 & 2.063 & 0.318 & 90.16 & \textbf{183.9} & 205.1 & 7.5610 \\
& SingleHDR + \textbf{Ours} & \textbf{64.82} & \textbf{60.34} & \textbf{12.17} & \textbf{11.17} & \textbf{4.056} & \textbf{4.419} & \textbf{2.151} & \textbf{0.301}  & \textbf{2.159} & \textbf{0.324} & \textbf{78.35} & 186.2 & \textbf{202.8} & \textbf{8.8604} \\
& GlowGAN~\cite{wang2023glowgan} & 79.02 & 77.49 & 14.54 & 14.25 & 4.008 & 4.209 & 1.890 & 0.342 & 1.900 & 0.346 & 93.79 & 256.7 & 275.4 & 6.0815 \\
& GlowGAN + \textbf{Ours} & \textbf{69.19} & \textbf{66.79} & \textbf{11.82} & \textbf{11.47} & \textbf{4.021} & \textbf{4.248} & \textbf{2.020} & \textbf{0.363} & \textbf{2.027} & \textbf{0.394} & \textbf{78.41} & \textbf{231.7} & \textbf{258.0} & \textbf{6.8441} \\
\cmidrule(lr){2-16}
& Deep Recursive HDRI~\cite{lee2018re} & 68.29 & 66.45 & 14.30 & 13.40 & 4.207 & 4.594 & 2.161 & 0.307 & 2.199 & 0.305 & 88.27 & 111.7 & 119.0 & \textbf{9.3884} \\
& Deep Recursive HDRI + \textbf{Ours} & \textbf{64.24} & \textbf{64.24} & \textbf{13.96} & \textbf{12.75} & \textbf{4.289} & \textbf{4.673} & \textbf{2.250} & \textbf{0.320} & \textbf{2.309} & \textbf{0.393} & \textbf{85.51} & \textbf{101.1} & \textbf{110.3} & 9.0358 \\
& Multi-Exp. Gen.~\cite{le2023single} & 71.51 & 65.40 & 13.89 & 12.71 & 3.900 & 4.410 & 2.013 & 0.209 & 1.964 & 0.262 & 66.12 & \textbf{228.3} & \textbf{275.8} & 7.8906 \\
& Multi-Exp. Gen. + \textbf{Ours} & \textbf{62.47} & \textbf{56.12} & \textbf{11.08} & \textbf{10.01} & \textbf{3.984} & \textbf{4.445} & \textbf{2.154} & \textbf{0.216} & \textbf{2.117} & \textbf{0.421} & \textbf{63.40} & 300.2 & 319.0 & \textbf{8.0233} \\
& CEVR~\cite{chen2023learning} & 72.42 & 70.21 & 14.33 & 13.48 & 4.044 & 4.421 & 2.043 & 0.299 & 2.093 & 0.299 & 90.44 & \textbf{123.4} & \textbf{135.3} & \textbf{9.2704} \\
& CEVR + \textbf{Ours} & \textbf{71.17} & \textbf{68.90} & \textbf{13.71} & \textbf{12.97} & \textbf{4.100} & \textbf{4.456} & \textbf{2.072} & \textbf{0.322} & \textbf{2.123}& \textbf{0.369} & \textbf{86.12} & 126.8 & 138.0 & 9.0358 \\
\midrule
\multirow{9}{*}{HDR-Eye} & SingleHDR~\cite{liu2020single} & 73.27 & 71.19 & 14.29 & 14.18 & 4.033 & 4.375 & 1.736 & 1.820 & 0.250 & 0.235 & 91.06 & 221.6 & 227.6 & \textbf{7.1277} \\
& SingleHDR + \textbf{Ours} & \textbf{62.50} & \textbf{57.72} & \textbf{12.02} & \textbf{11.12} & \textbf{4.087} & \textbf{4.446} & \textbf{1.875} & \textbf{1.955} & \textbf{0.251} & 0.235 & \textbf{78.39} & \textbf{213.6} & \textbf{227.0} & 6.8385 \\
& GlowGAN~\cite{wang2023glowgan} & 70.07 & 70.51 & 14.14 & 13.92 & 4.211 & 4.346 & 1.873 & 1.886 & 0.259 & 0.259 & 89.48 & 226.5 & 245.2 & \textbf{6.4705} \\
& GlowGAN + \textbf{Ours} & \textbf{61.00} & \textbf{61.67} & \textbf{11.98} & \textbf{11.75} & \textbf{4.277} & \textbf{4.460} & \textbf{1.972} & \textbf{2.011} & \textbf{0.260} & \textbf{0.260} & \textbf{79.29} & \textbf{181.3} & \textbf{225.1} & 6.4286 \\
\cmidrule(lr){2-16}
& Deep Recursive HDRI~\cite{lee2018re} & 65.61 & 64.98 & 13.21 & 11.70 & 4.225 & 4.620 & 1.900 & 1.933 & 0.185 & 0.176 & 87.68 & \textbf{199.1} & 200.5 & \textbf{7.2791} \\
& Deep Recursive HDRI + \textbf{Ours} & \textbf{65.04} & \textbf{62.95} & \textbf{12.21} & \textbf{10.94} & \textbf{4.260} & \textbf{4.674} & \textbf{1.950} & \textbf{2.019} & \textbf{0.190} & \textbf{0.180} & \textbf{84.55} & 204.8 & \textbf{192.6} & 7.0809 \\
& Multi-Exp. Gen.~\cite{le2023single} & 66.71 & 61.07 & 13.38 & 11.53 & 4.000 & 4.539 & 1.791 & 1.859 & 0.158 & 0.158 & 67.98 & \textbf{229.0} & \textbf{228.9} & \textbf{6.1145} \\
& Multi-Exp. Gen. + \textbf{Ours} & \textbf{62.74} & \textbf{56.14} & \textbf{11.06} & \textbf{9.80} & \textbf{4.113} & \textbf{4.559} & \textbf{2.075} & \textbf{2.106} & \textbf{0.186} & \textbf{0.199} & \textbf{67.44} & 338.8 & 302.0 & 5.8216 \\
& CEVR~\cite{chen2023learning} & 69.31 & 68.04 & 14.69 & 13.57 & 4.082 & 4.447 & 1.792 & 1.823 & 0.197 & 0.174 & 89.64 & \textbf{202.8} & \textbf{201.2} & \textbf{7.2253} \\
& CEVR + \textbf{Ours} & \textbf{68.25} & \textbf{66.16} & \textbf{12.46} & \textbf{11.29} & \textbf{4.100} & \textbf{4.478} & 1.792 & \textbf{1.851} & \textbf{0.205} & \textbf{0.183} & \textbf{83.80} & 219.9 & 208.6 & 7.0871 \\
\bottomrule
\end{tabular}
}
\end{table*}
\subsection{Experimental Setup}

\paragraph{Implementation Details.}

We use SDXL-base1.0 with an off-the-shelf depth-conditioned ControlNet~\cite{zhang2023adding} as the diffusion inpainting backbone model. 
\red{In this work, we use a fixed and generic set of parameters and input prompts for evaluating all the cases.}
During inference, we use $50$ sampling steps, a $5.0$ guide scale, and a $0.5$ ControlNet conditioning scale. \blue{Additionally, we set the saturation threshold to 245.}
For SDEdit strength scheduling, we use a linear decay schedule, starting at 0.95 and decreasing by 0.05 per iteration, \ie, [0.95, 0.90, 0.85, 0.80, ...]. \red{The compensation scale has a linear increasing schedule, starting at 0.2 and increasing by 0.1 per iteration, \ie, [0.2, 0.3, 0.4, 0.5, ...].}

\red{To avoid a single specially chosen prompt from disproportionately boosting performance, we use a fixed, generic text prompt ``{\em A beautiful photo with bright background. High-resolution image with a lot of details and sharpness. 4K, Ultra Quality. Good photo.}'' to minimize the sensitivity of generative models to prompt variations.}

\vspace{-3mm}
\paragraph{Baseline Methods.}
We use four indirect HDR reconstruction methods: One is the simple gamma 2.2 tone mapping curve, and the others are existing indirect HDR reconstruction methods, including CEVR \cite{chen2023learning}, Multi-Exposure Generation \cite{le2023single}, and Deep Recursive HDRI \cite{lee2018deep}. 
We also use two direct methods, including GlowGAN \cite{wang2023glowgan} and singleHDR \cite{liu2020single}, for comprehensive evaluations.


\vspace{-3mm}
\paragraph{Datasets.}
We conduct the experiments on two public HDR datasets, including the VDS \cite{lee2018deep} and HDR-Eye \cite{nemoto2015visual} datasets. 
Both of them cover outdoor and indoor scenes. 


\vspace{-3mm}
\paragraph{Evaluation Metrics.}
To comprehensively evaluate the performance of our method, we use the full-reference metrics HDR-VDP-3 \cite{Mantiuk2023HDRVDP3} and KID score \cite{bińkowski2018demystifying}, which assess the fidelity of generation. 
In addition, we use the non-reference metrics BRISQUE \cite{Anish2012brisque}, NIQE \cite{Anish2012niqe}, NIMA \cite{Hossein2018nima}, MUSIQ \cite{ke2021MUSIQ}, CLIP-IQA \cite{wang2022exploring}, and PU21-PIQE \cite{Mantiuk2021pu21} for the overall image quality evaluation.


\vspace{-3mm}
\paragraph{Tone Mapping Operators.}
The majority of image quality metrics are not compatible with HDR file inputs. 
Therefore, we perform tone mapping to convert the HDR results to LDR prior to comparison, ensuring that they share the same dynamic range.


For this process, we employ two different tone-mapping operators (TMOs): Reinhard’s (RH's) method \cite{Reinhard2002Photographic} and Kim and Kautz’s (KK's) method \cite{Kim2008Consistent}. 
These operators compress the full HDR dynamic range into a more limited range while preserving details without clipping.


\begin{figure*}[t]
\centering
\includegraphics[width=1.0\textwidth]{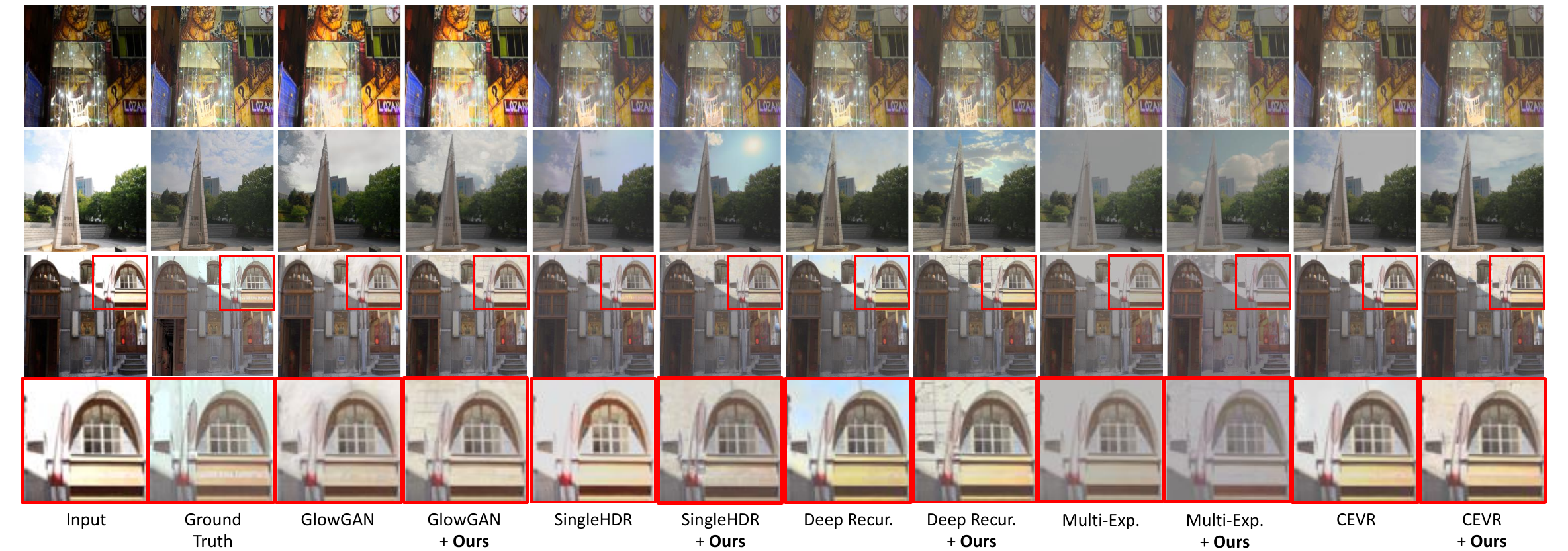}
\vspace{-6mm}
\caption{
\textbf{Qualitative comparisons of HDR reconstruction methods.} Our approach enhances various baseline methods (indicated by ``+ \textbf{Ours}'') across diverse scenes, including outdoor architecture, indoor lighting, and urban landscapes. While existing methods often produce unrealistic content or artifacts in over-exposed regions, our method generates more natural details and lighting patterns while maintaining structural consistency without training. Results show improved reconstruction of sky details, cloud patterns, and lighting effects across different baseline methods. The top row shows the main difference between the baseline results and hallucinated results. Though we can generate a similar texture and color tone, the generated content and ground truth might not be considered the same by the KID score and HDR-VDP-3. 
}
\label{fig:visual}
\end{figure*}

\subsection{Results}
\paragraph{Quantitative Comparisons.}
Our method demonstrates notably superior performance in non-reference metrics and consistently outperforms the baseline methods across both VDS and HDR-Eye datasets, as reported in Tab.~\ref{tab:baseline_ours}. 
While we observe slightly lower scores in reference-based metrics like KID and HDR-VDP-3, this is expected since our method focuses on plausible reconstruction of over-exposed regions rather than strict adherence to the ground truth. 
Standard reference-based metrics inherently penalize creative reconstruction of lost information, even when the results are perceptually compelling. 
%
%
These results substantiate the effectiveness of our inpainting pipeline in reconstructing challenging areas where traditional methods struggle.


\begin{figure}[t]
\centering
\includegraphics[width=1.0\columnwidth]{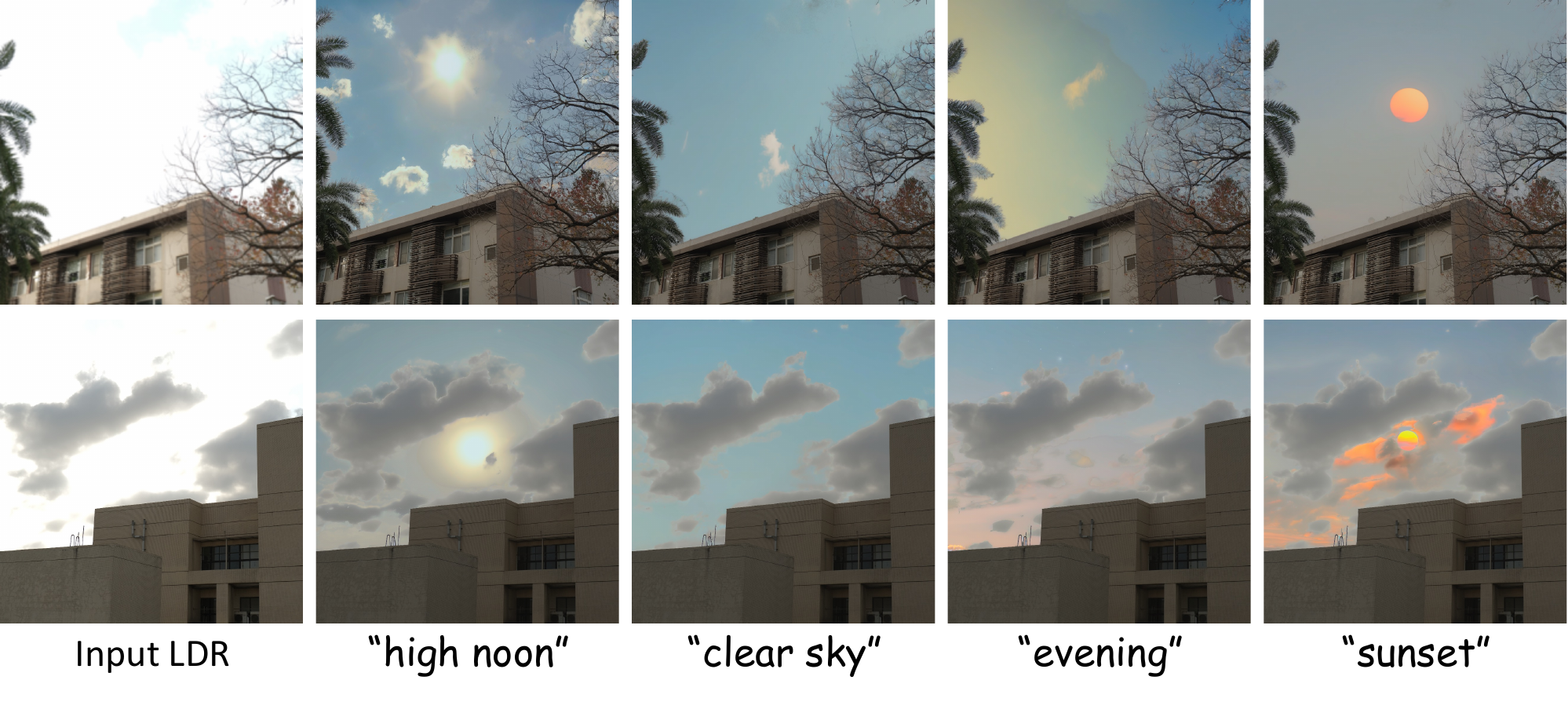}
\vspace{-3mm}
\caption{
\textbf{Inpainting diversity results.} By leveraging the diffusion prior, our inpainting pipeline gains the ability to generate diverse scenes with different lighting conditions and sky appearances from the same input LDR image. 
}
\label{fig:diversity}
\end{figure}

\vspace{-3mm}
\paragraph{Qualitative Comparisons.}
With our diffusion inpainting pipeline, we can generate natural and diverse over-exposed scenes for the HDR images. 
Qualitative results, as displayed in Fig.~\ref{fig:visual}, illustrate the differences between the baseline methods and the boosted baseline results one by one.  
In addition, we compare our results to those by GlowGAN. 
Our method leverages the diffusion prior and can produce more feasible HDR scenes than GlowGAN. 
We also explore the impact of different text prompts. 
As visualized in Fig. \ref{fig:diversity}, our method exhibits adaptability to all the text prompts and hallucinates plausible results. 
More examples of in-the-wild or extreme cases can be found in supplementary materials.


\begin{table}[t]
\caption{\blue{\textbf{Ablation study on diffusion backbones.} Our algorithm improves state-of-the-art methods on the VDS dataset~\cite{lee2018deep} with different diffusion backbones without additional training.}}
\label{tab:ablation}
\vspace{-0.9pc}
    \centering
    \scriptsize
    \label{tab:ablation}
    \resizebox{1.0\columnwidth}{!} {

\begin{tabular}{l*{7}{c}}
\toprule
\cmidrule(lr){2-6} 
Method & \multicolumn{2}{c}{NIQE ↓} & \multicolumn{2}{c}{NIMA ↑} & PU21-PIQE ↓ \\
\cmidrule(lr){2-3} \cmidrule(lr){4-5}  
\multicolumn{1}{r}{TMO} & RH's & KK's & RH's & KK's & -- \\ 

\midrule
CEVR~\cite{chen2023learning} & 14.33 & 13.48 & 4.044 & 4.421 & 90.44 \\

CEVR + \textbf{Ours (SDXL)} & 13.71 & 12.97 & 4.100 & 4.456 & 86.12 \\

CEVR + \textbf{Ours (SD turbo)} & \textbf{12.36} & \textbf{11.60} & \textbf{4.150} & \textbf{4.482} & \textbf{78.46} \\ 

\bottomrule
    \end{tabular}
    }
\vspace{-1pc}
\end{table}

\begin{figure}[t]
\centering
\includegraphics[width=1.0\columnwidth]{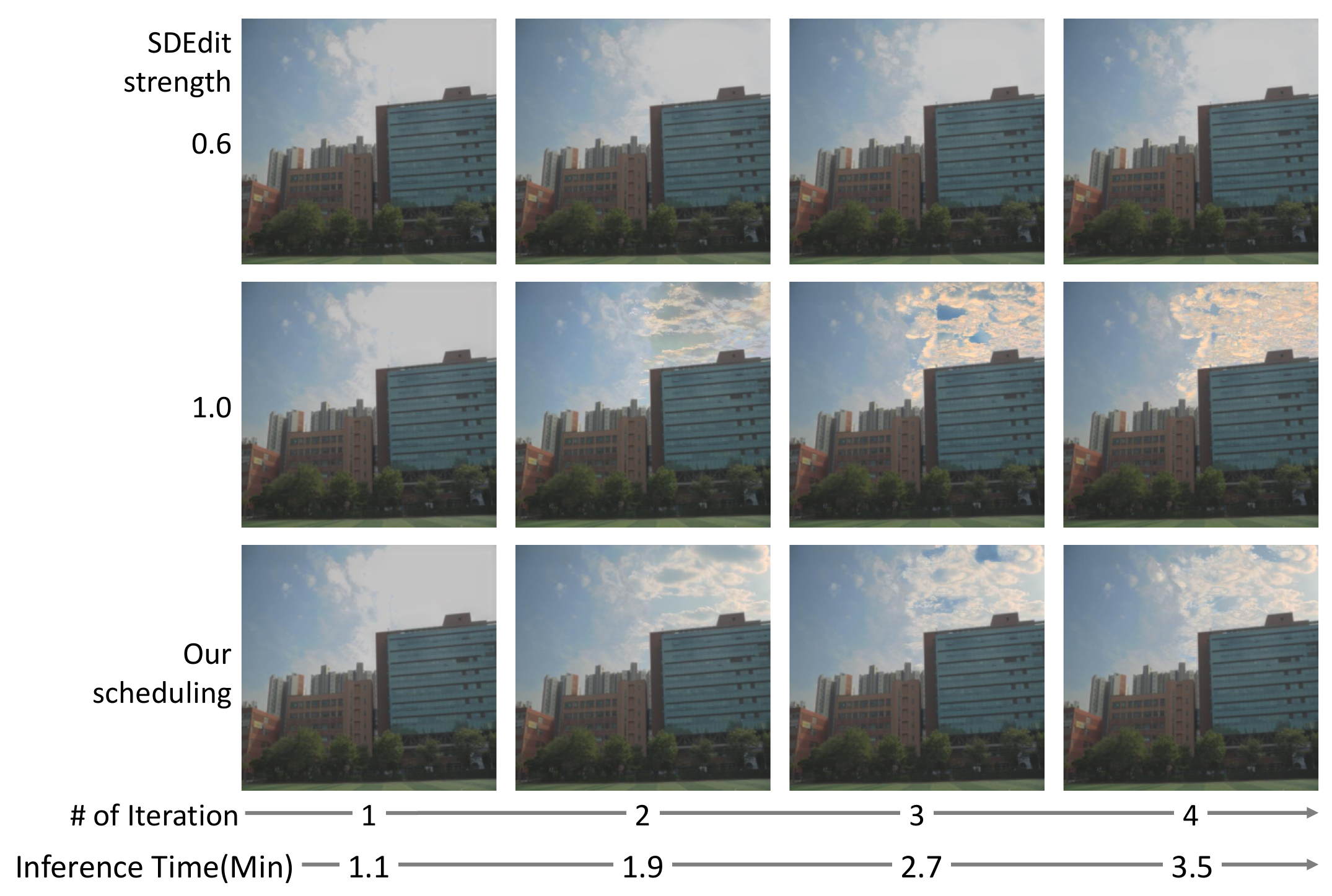}
\vspace{-6mm}
\caption{
\textbf{SDEdit strength scheduling ablation study.} Low strength, \eg, $0.6$, limits content generation in over-exposed regions, while high strength, \eg, $1.0$, causes inconsistency between iterations. Our scheduling gradually decreases noise strength to balance content generation with detail refinement and consistency across all the four iterations.
}
\label{fig:ablation_sdedit}
\end{figure}

\subsection{Ablation Studies}
\paragraph{SDEdit Strength Scheduling.}
In the inpainting pipeline, we use varying SDEdit strengths to control the balance of creativity and consistency of texture between iterations. 
Fig.~\ref{fig:ablation_sdedit} shows the results by using different SDEdit strengths. 
When the strength is fixed to a lower value, \eg, $0.6$, the results of the next iteration would lose the generative diversity and be the same as those in the previous iteration. 
In contrast, a fixed high strength of $1.0$ prevents the inpainting model from taking the results of the previous iteration into consideration, resulting in the production of infeasible images.


\begin{figure}[t]
\centering
\includegraphics[width=1.0\columnwidth]{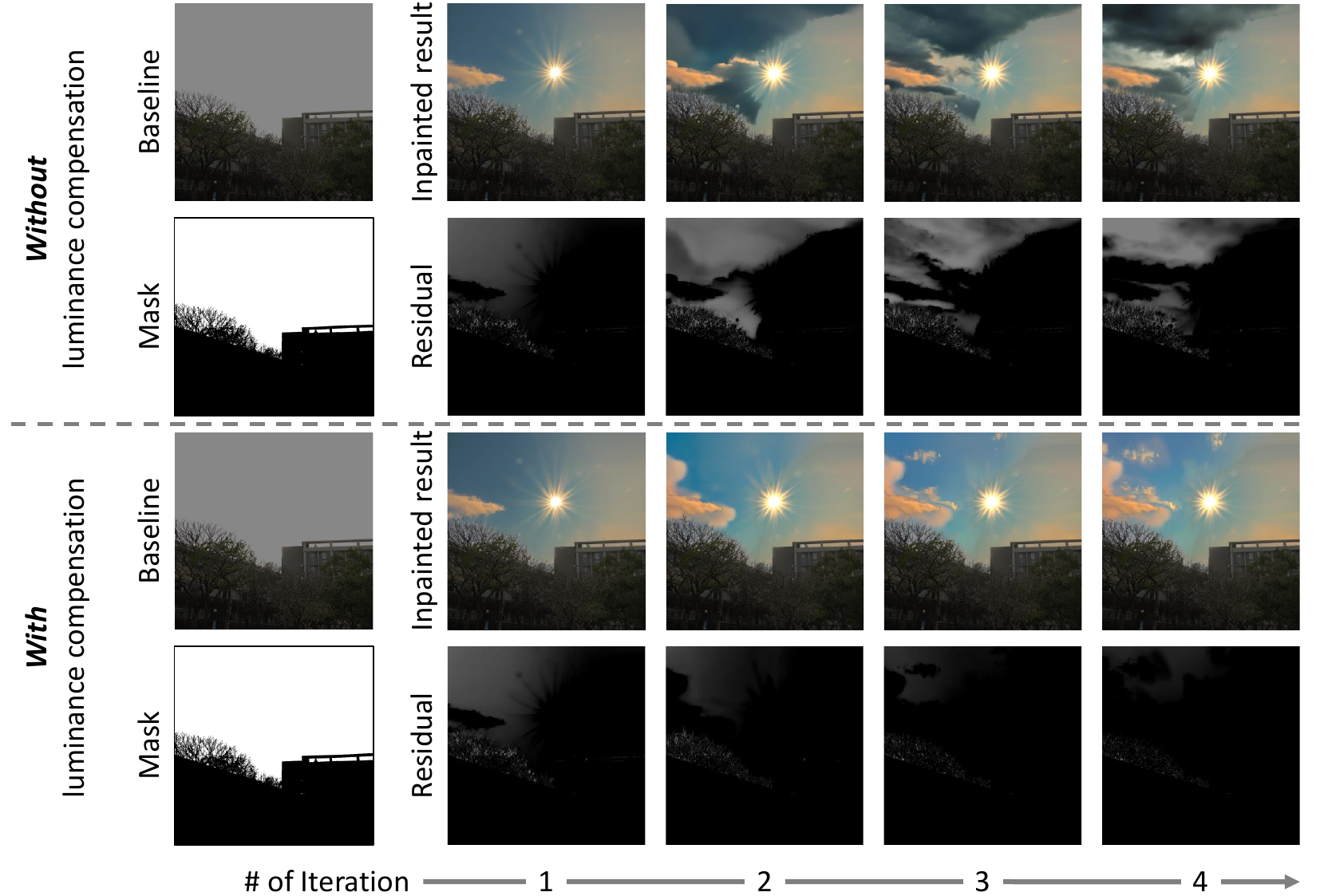}
\vspace{-6mm}
\caption{
\textbf{Luminance compensation ablation study.} 
(\emph{Top}) Without compensation, the inpainted content may be unrealistically dark. (\emph{Bottom}) Our compensation adjusts pixel values to maintain physically plausible brightness levels, as shown by decreasing residuals across iterations.
}
\label{fig:ablation_compensation}
\end{figure}

\vspace{-4mm}
\paragraph{Compensation Block.}
When generating details for overexposed regions, we aim to prevent the generation process from modifying the original input LDR in a way that leads to an unrealistic inverse CRF. 
To ensure this, we constrain the baseline methods by setting their pixel values to the lower-bound of the overexposed areas. 
If inpainted pixels fall below this lower-bound, we correct them by adding the absolute difference between the inpainted value and the lower-bound, scaled by a compensation factor. Additionally, we refine the mask to better align with the adjusted pixel values. 
Fig.~\ref{fig:ablation_compensation} shows the results with and without the compensation block. 
As demonstrated in the second row, we find that the residual fails to diminish without the compensation block as the iteration progresses.
In the fourth row, it is evident that the residual is significantly reduced with the aid of this compensation block. \blue{The quantitative evaluation of each component of our approach is reported in Tab. \ref{tab:ablation_quan}.}


\begin{table}[t]

    \centering
    \scriptsize
    \caption{\blue{\textbf{Quantitative ablation studies.} We conducted a quantitative evaluation on the VDS dataset and demonstrated that our approach yields improvements in non-reference image quality metrics.}}
    \label{tab:ablation_quan}
    \vspace{-3mm}
    \resizebox{1.0\columnwidth}{!} {
    \begin{tabular}{l|ccc}
    \toprule
    Components (RH's TMO)  & NIQE $\downarrow$ & NIMA $\uparrow$ & PU21-PIQE $\downarrow$ \\
    \midrule
    Baseline (CEVR)               & 14.33 & 4.044  & 90.44  \\
    w/o Merge \& Comp.     & 13.57 & 4.045  & 85.15  \\
    w/o Merge              & 13.60 & 4.021  & \textbf{84.68}  \\
    w/o Comp.                    & \textbf{13.39} & 4.028  & 85.07  \\
    \textbf{Ours}                & 13.71 & \textbf{4.100} & 86.12  \\
    \bottomrule
    \end{tabular}
    }
\end{table}

\vspace{-5mm}
\paragraph{Diffusion Backbone.}
The training-free nature of our method ensures that the inpainting pipeline exhibits consistent performance across various diffusion backbone models. 
To demonstrate this, we conduct experiments using two distinct diffusion models with depth-conditioned ControlNet, including SDXL, the original model employed in our method, and SDXL Turbo, a distilled variant of SDXL characterized by an accelerated inference speed but diminished image quality and prompt alignment.
As presented in Tab.~\ref{tab:ablation}, our method with SDXL turbo still beats the baseline method in all non-reference-based metrics. 
Notably, it even surpasses the performance with SDXL in certain metrics.


\vspace{-3mm}
\paragraph{Inference Time.} 
We conduct the evaluations with different numbers of iterations on the VDS dataset, using the output image size of \emph{$1024 \times 1024$}.
Fig.~\ref{fig:ablation_sdedit} shows the computation time with different numbers of iterations. We conducted this analysis with single RTX 3090.


\subsection{Limitation}
\begin{figure}[t]
\centering
\includegraphics[width=1.0\columnwidth]{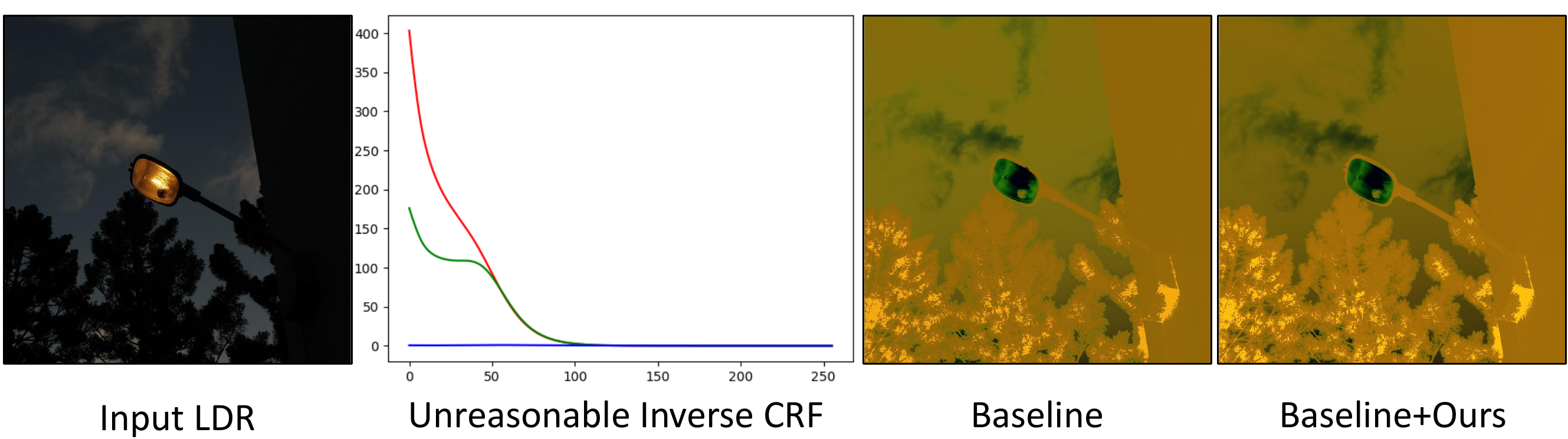}
\vspace{-6mm}
\caption{
\textbf{Limitation.} 
Our method is restricted by the estimated inverse CRF. When the estimated inverse CRF is unreasonable, our results suffer from color shifting as well.
}
\label{fig:limitation}
\end{figure}

Although our method effectively boosts the baseline methods by iteratively inpainting the LDR stacks, its performance is still contingent upon the baseline method producing a reasonable inverse CRF. 
As shown in Fig.~\ref{fig:limitation}, the baseline method yields a non-monotonic inverse CRF. The non-monotonic inverse CRF disrupts the assumed one-to-one relationship between pixel intensity and scene radiance. 
Thus, both the baseline and the inpainted results suffer from color-shifting side effects. 


\section{Conclusion}
We present a training-free, diffusion-based pipeline for HDR restoration of over-exposed regions. Our method leverages diffusion priors with SDEdit refinement to generate plausible content while preserving exposure consistency. Through iterative refinement and compensation mechanisms, our method consistently and substantially enhances existing HDR reconstruction methods, particularly in over-exposed areas. Comprehensive results demonstrate significant improvements across metrics and datasets, highlighting the effectiveness of our approach across diverse scenes without the need for additional training.
\blue{As a promising direction for future work, we plan to extend our framework to handle under-exposed regions, aiming for a unified solution that addresses both extremes of the dynamic range.}

\paragraph{Acknowledgments.} This work was supported in part by National Science and Technology Council (NSTC) under grants 111-2628-E-A49-025-MY3, 112-2221-E-A49-090-MY3, 111-2634-F-002-023, 111-2634-F-006-012, 110-2221-E-A49-065-MY3, 111-2634-F-A49-010, 112-2222-E-A49-004-MY2, and 113-2628-E-A49-023-. This work was funded in part by MediaTek. Yu-Lun Liu acknowledges the Yushan Young Fellow Program by the MOE in Taiwan.
{
    \small
    \bibliographystyle{ieeenat_fullname}
    \bibliography{main}
}

\end{document}